\documentclass[letterpaper, 10 pt, conference]{ieeeconf}  

\IEEEoverridecommandlockouts                              

\usepackage{graphics} 
\usepackage{epsfig} 
\usepackage{mathptmx} 
\usepackage{times} 
\usepackage{amsmath} 
\usepackage{amssymb}  
\usepackage[at]{easylist}
\usepackage{mathtools}
\usepackage{setspace}
\usepackage{soul}
\usepackage{color}
\usepackage[binary-units=true]{siunitx}
\usepackage{xcolor}
\usepackage{mathtools}
\usepackage{float}
\usepackage[ruled,vlined,linesnumbered,noresetcount]{algorithm2e}
\usepackage[pagebackref=true,breaklinks=true,colorlinks,bookmarks=false]{hyperref}

\title{\LARGE \bf Object manipulation through contact configuration regulation: \\ multiple and intermittent contacts}
\author{
Orion Taylor$^{1}$ \quad Neel Doshi$^{2}$ \quad Alberto Rodriguez\vspace{0.2cm}\\
Massachusetts Institute of Technology\vspace{0.2cm}\\
\vspace{-0.4cm}
\thanks{
\hspace{-0.4cm}
$^1$ This work was supported by an ARA/Sponsored research award from Amazon. \newline
$^2$ Neel Doshi (\texttt{ndd@amazon.com}) is with Amazon Robotics R\&D. This research was conducted prior to Neel joining Amazon.
}
}

\def\*#1{\mathbf{#1}}
\def\?#1{\mathbb{#1}}

\newcommand{\secref}[1]{Section~\ref{#1}}

\newcommand{\figref}[1]{Fig.~\ref{#1}}

\newcommand{\myparagraph}[1]{\vspace{0.05in}\noindent\textbf{#1}}

\begin{document}

\maketitle

\thispagestyle{empty}
\pagestyle{empty}

\begin{abstract}

In this work, we build on our method for manipulating unknown objects via \textit{contact configuration regulation}: the estimation and control of the location, geometry, and mode of all contacts between the robot, object, and environment. We further develop our estimator and controller to enable manipulation through more complex contact interactions, including intermittent contact between the robot/object, and multiple contacts between the object/environment. In addition, we support a larger set of contact geometries at each interface. This is accomplished through a factor graph based estimation framework that reasons about the complementary kinematic and wrench constraints of contact to predict the current contact configuration. We are aided by the incorporation of a limited amount of visual feedback; which when combined with the available F/T sensing and robot proprioception, allows us to differentiate contact modes that were previously indistinguishable. We implement this revamped framework on our manipulation platform, and demonstrate that it allows the robot to perform a wider set of manipulation tasks. This includes, using a wall as a support to re-orient an object, or regulating the contact geometry between the object and the ground. Finally, we conduct ablation studies to understand the contributions from visual and tactile feedback in our manipulation framework. Our code can be found at: \url{https://github.com/mcubelab/pbal}.

\end{abstract}

\section{Introduction}
\label{sec:intro}

Robotic manipulation is driven by contact: robots manipulate objects by imparting wrenches and motions through the contact interfaces of the system. The geometric and frictional properties of these contacts determine which wrenches and motions are feasible. As such, regulating an object's \textit{contact configuration} -- the location, mode and geometry of all contacts between the object, robot, and environment -- is a fundamental aspect of manipulation. For example, when folding origami or tying one's shoes, regulating sticking contact between the hand and the paper or shoe-lace ensures that the latter moves as intended. 

\begin{figure}
    \centering
    \includegraphics[width=0.9\columnwidth]{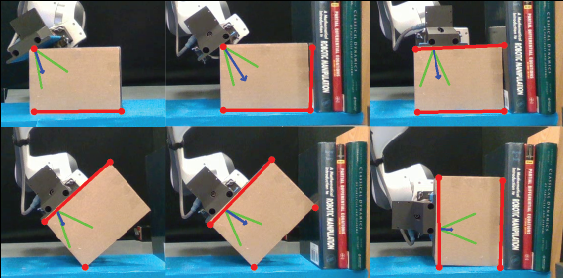}
    \vspace{-0.35cm}
    \caption{Manipulating a box by first pushing it against a horizontal wall, and then using the wall as a support to reorient the box. Executing this involves several different combinations of contact geometries between the object and the environment/hand. In our framework, the measured robot wrench (blue) and the robot pose are used to estimate the friction constraints (green) and the contact locations and geometries (red). }
    \label{fig:introduction_figure}
    \vspace{-0.5cm}
\end{figure}

In our previous work \cite{previous_paper}, we developed a joint estimation and control framework for manipulating 2D objects in the gravity plane via contact configuration regulation. The key estimation challenge was inferring the properties of the object/environment contact(s) from measurements taken at the object/robot contact(s), without prior knowledge of the object's geometry. Previously, we restricted the set of possible contact geometries that were considered to prevent ambiguity. While effective, this assumption constrained the capabilities of the robot. This work pushes beyond these limitations by expanding our framework to include a more diverse set contact interactions between the robot/object/environment. 

Consider, for example, the common warehouse task of loading (or unloading) boxes from a storage shelf (\figref{fig:introduction_figure}). To accomplish this, it is natural to reason about the different possible combinations of contact formations between the box and the robot/environment. With the above in mind, the key contribution of our research is the development of a manipulation system that utilizes the mechanics of contact (\secref{sec:system_overview}) to generate dexterous behaviors with unknown objects (\secref{sec:demonstrations}) through a variety of contact formations. In particular, we: 

\begin{itemize}
\item Present an improved estimation framework that uses factor graphs and limited visual feedback in order to localize, characterize, and differentiate a more diverse set of contact configurations (\secref{sec:estimation_framework}).
\item Use these improved estimates to develop new manipulation primitives for our controller to regulate the system through these additional contact types, including manipulating an object against a corner (\secref{sec:control_framework}).
\end{itemize}

We conclude that, for our purposes, the unique strength of vision is in revealing the global geometry/topology of the system, as opposed to the dense local information provided by F/T sensing and proprioception. Even coarse, low-frequency vision measurements can significantly magnify the inferential power of the available high-quality tactile sensing, allowing us to disambiguate contact configurations that would be indistinguishable when relying on feedback from a single source (\secref{sec:experiments}). We demonstrate that this allows us to perform more complex manipulation tasks while still being agnostic to object geometry (\secref{sec:demonstrations}).

\begin{figure}
    \centering
    \includegraphics[width=0.9\columnwidth]{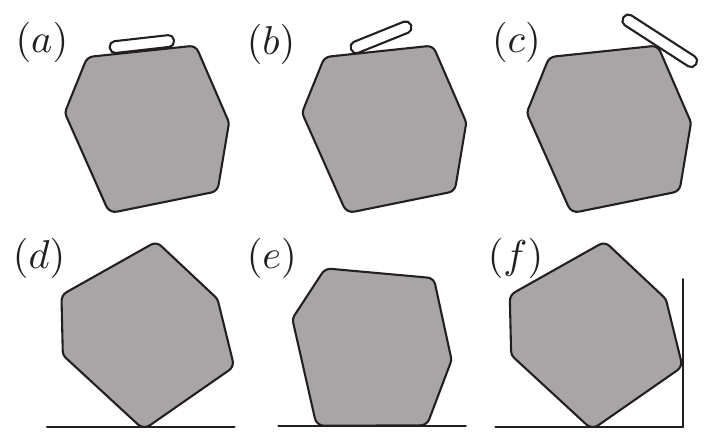}
    \vspace{-0.35cm}
    \caption{Different contact geometries between the robot hand/object (top) and object/environment (bottom) supported by our system.}
    \label{fig:contact_modes_figure}
\end{figure}

\section{Related work}
\label{sec:lit_review}

Prior work on contact configuration regulation has often focused on tasks such as polishing \cite{FURUKAWA1996261} and deburring \cite{her1991automated}, as well as opening a variety of drawers \cite{karayiannidis2013model} or doors \cite{niemeyer1997simple}.
The ability to directly observe all contacts facilitates estimation and control. 
Less attention has been given to manipulation tasks where
not all contacts are directly observable (e.g., during non-prehensile manipulation).
In this area, prior research usually either focuses on control assuming a known model of the world~\cite{hogan2020reactive, Hou-RSS-20, hogan2020tactile}, or estimation assuming stable interactions~\cite{ma2021icra, shirai2023tool}.
While there is more recent work on joint estimation and control, individual papers often make simplifying assumptions (e.g., frictionless models of contact) \cite{Lefebvre2003polyhedral, de2007constraint}, or learn task-specific policies from data (e.g, for cable manipulation \cite{she2021cable}, part insertion \cite{dong2021icra}, or manipulating rigid objects on a shelf \cite{liang2022HFVC}). In contrast, our contribution is an object-agnostic joint estimation and control framework that reasons about all frictional interactions between the robot, object, and environment, leveraging limited visual feedback to execute more complex manipulations than in our previous work 
\cite{previous_paper}. 



\myparagraph{Estimation} The primary focus of this work is the fusion of vision and tactile information for improved estimation of the location, geometry, and mode of all contacts between the robot, object, and environment. There is a large body of work in the area of localizing contacts/objects and estimating contact configurations using vision and/or tactile feedback. Many of these works present solutions to specific aspects of the problem we tackle here. Prior information about the end-effector geometry can be exploited to localize contacts \cite{bicchi1993contact, manuelli2016localizing, yu2018realtime, wang2020contact}. Similarly, knowledge of the object geometry can be used estimate its pose using only tactile sensing \cite{petrovskaya2011touch}, or a fusion of both visual and tactile feedback \cite{bimbo2013vision, Hebert2011sensorfusion}. 

Particle filters are a popular formulation for nonlinear estimation problems, and have been often used in the context of tactile localization. Koval et al. \cite{koval2015poseestimation} develop an approach to particle filters that takes into account how contact restricts an object's degrees-of-freedom. Li et al. \cite{li2015intermittent} explicitly reason about the complementarity constraints of frictional contact, while Meeussen et al. \cite{Meeussen2006sensorfusion} apply particle filters to fuse visual and tactile data. These works partially overlap with the problem that we are trying to solve.


A core part of our estimation framework is also inferring the contact wrench constraints (i.e., the generalized friction cone \cite{erdmann1994representation}). Prior work in this field has focused on the problem of planar pushing \cite{yoshikawa1991identification, lynch1993estimating, zhou2018convex}. Of particular relevance is the work of Zhou et al. \cite{zhou2018convex}, who estimate the set of wrenches that can be transmitted through planar frictional contact (i.e., the limit-surface \cite{goyal1991planar}). 

Finally, several approaches have been presented to the problem of joint shape and pose estimation for 2D objects \cite{strub2014hapticshape, suddhu2021SLAM, peter2018SLAM}. Suresh et al. \cite{suddhu2021SLAM} use factor graphs \cite{ISAM2,gtsam} to encode the kinematic constraints of contact, a method that we also employ. However, our work places a significant emphasis on estimating \textit{extrinsic contact formations} \cite{daolin2021extrinsic} (i.e., between the object and environment) that are not directly measured. 



\section{System Overview}
\label{sec:system_overview}

\begin{figure}
    \centering
    \includegraphics[width=0.9\columnwidth]{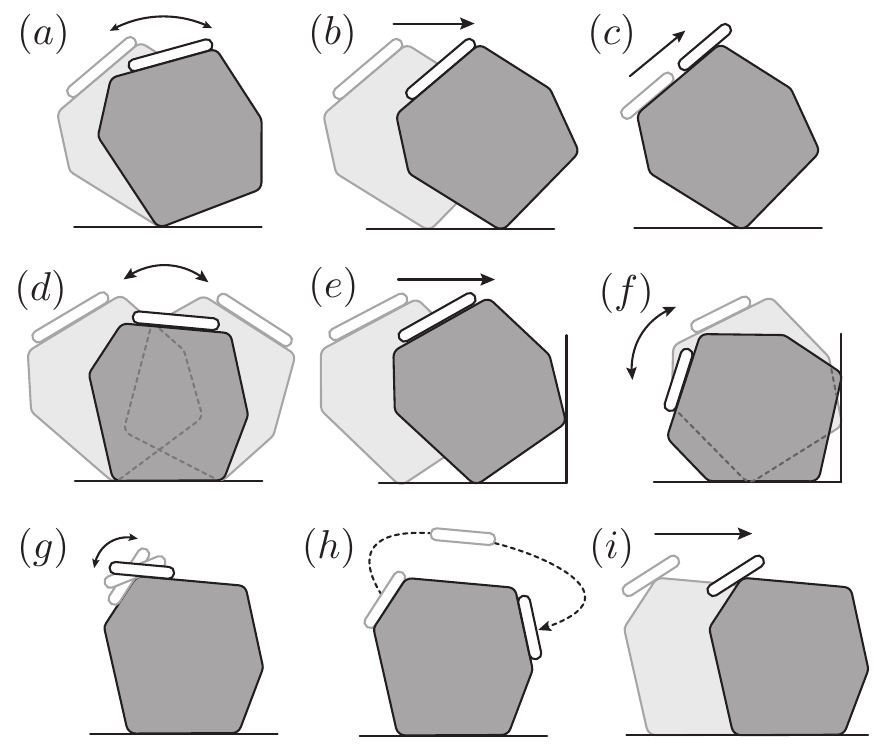}
    \vspace{-0.35cm}
    \caption{Contact configuration transitions that can occur in our system.}
    \label{fig:contact_transitions_figure}
    \vspace{-0.5cm}
\end{figure}

We focus on quasi-static manipulation of objects on top of a horizontal work surface. The system consists of: the robot \textbf{hand}, which we represent as a line of length $2l_h$, the \textbf{object}, which we treat as planar convex polygon, and the \textbf{environment}, which features the ground (a fixed horizontal line), and up to two vertical walls.

In our prior work, we assumed that the object was always in flush contact with the hand (\figref{fig:contact_modes_figure}a), the object was always in single point contact with the ground (\figref{fig:contact_modes_figure}d), and that the object's respective contact face and point at these two interfaces were fixed. These restrictions greatly simplified the system model, while still allowing for a contact-rich set of behaviors, primarily: pivoting the object about its ground contact, (\figref{fig:contact_transitions_figure}a), translating the object (\figref{fig:contact_transitions_figure}b), and sliding the hand relative to the object (\figref{fig:contact_transitions_figure}c).

Here we build on this framework, relaxing some of these assumptions to regulate more complex geometric interactions between the object and the hand/environment. We expand the set of admissible contact interactions to include:
\begin{itemize}
\item Object-line/hand-point contact (\figref{fig:contact_modes_figure}b), and object-point/hand-line contact (\figref{fig:contact_modes_figure}c).
\item Flush ground contact (\figref{fig:contact_modes_figure}e), and other external contacts, e.g., a wall (\figref{fig:contact_modes_figure}f). 
\item Transitions between contact geometries (\figref{fig:contact_transitions_figure}d-i). This could be simple, like the object making and breaking contact with a wall (\figref{fig:contact_transitions_figure}e); or more complex, like the hand transitioning from flush contact with one object face to the next by pivoting across a vertex (\figref{fig:contact_transitions_figure}g).
\end{itemize}

\subsection{System Parameterization}
\label{subsec:system_parameterization}

\begin{figure}
    \centering
    \includegraphics[width=0.9\columnwidth]{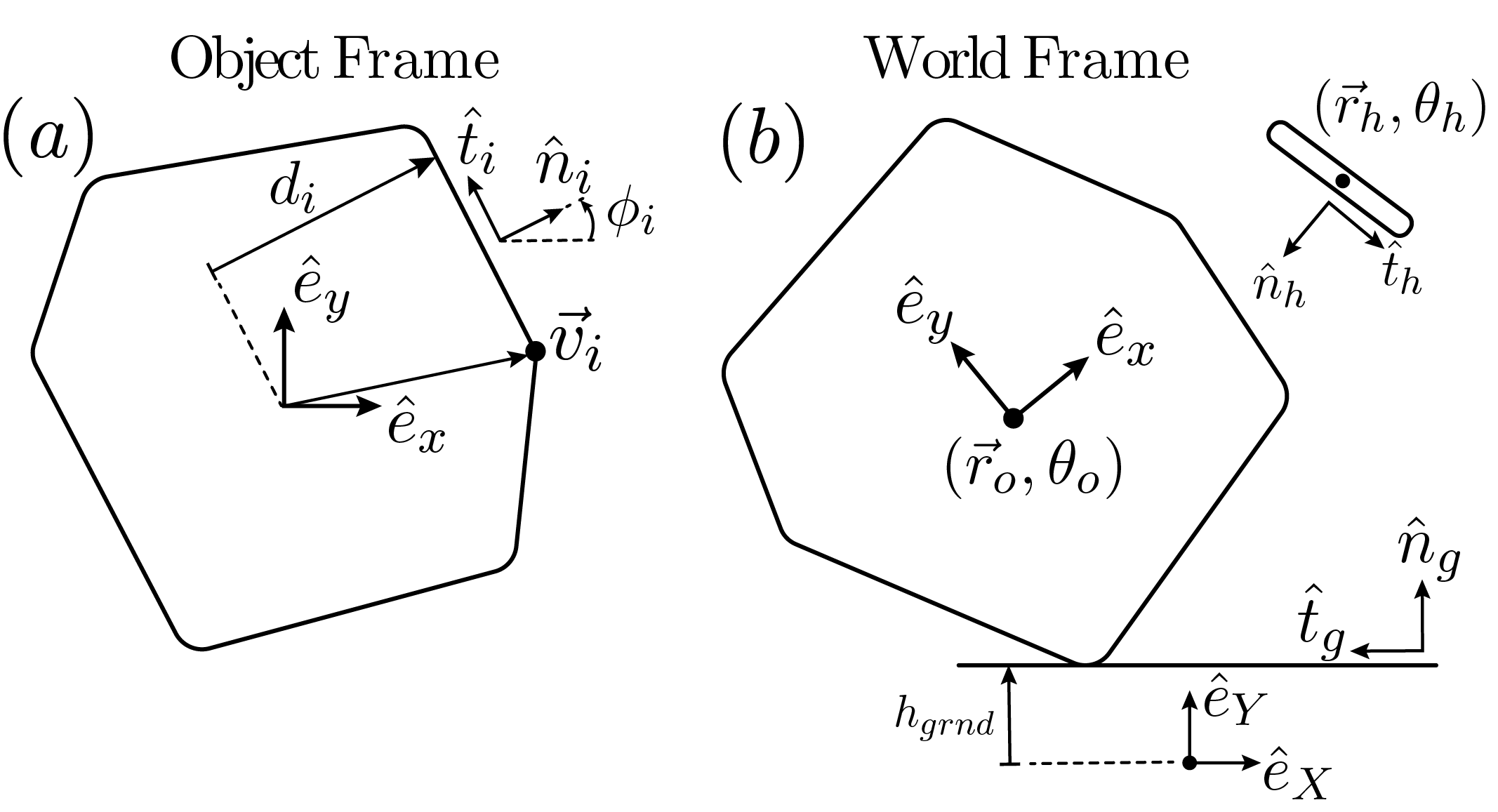}
    \vspace{-0.35cm}
    \caption{System parameterization.}
    \label{fig:parameterization_figure}
    \vspace{-0.5cm}
\end{figure}

The system's state consists of the planar poses of the hand $\vec{x}_h =(\vec{r}_h, \theta_h)$ and object $\vec{x}_o = (\vec{r}_o, \theta_o)$ in the world-frame (\figref{fig:parameterization_figure}b). The object, which is treated as a convex polygon, is parameterized by its (object-frame) vertex positions $\vec{v}_i = (x_i,y_i)$, outward facing surface normals $\hat{n}_i = (\cos\phi_i,\sin\phi_i)$, and contact face offsets $d_i =\hat{n}_i \cdot \vec{v}_i$ (\figref{fig:parameterization_figure}a). We denote the (world-frame) contact normals and contact tangents of the hand and ground as $(\hat{n}_h,\hat{t}_h)$ and $(\hat{n}_g,\hat{t}_g)$ respectively.  The constant $h_{grnd}$ represents the height of the ground. During flush contact with the hand, we also consider the relative tangential displacement between the hand and the object, $s$.

When the object is in single point contact with the ground, we model the effect of gravity  using the parameters  $(\alpha_i,\beta_i)$, which encapsulate the object's weight $(mg)$, the length $(l_i)$ of the gravitational moment arm (the vector from the ground contact point to the object's COM), and the angle $(\theta_o +\psi_i)$ of the gravitational moment arm:
\begin{align}
mgl_i\sin(\theta_o+\psi_i)=\alpha_i\cos(\theta_o)+\beta_i\sin(\theta_o)
\end{align}
From the quasi-static motion assumption, it follows that:
\begin{align}
    \label{eq:force_static_eq}
    \sum \vec{w}_{net} = \vec{w}_{h} + \vec{w}_{e} + \vec{w}_{grav} = 0,
\end{align}
where $\vec{w}_{h} = (\vec{F}_h,\tau_h)$, $\vec{w}_e = (\vec{F}_e,\tau_e)$, and $\vec{w}_{grav}=(-mg\hat{n}_g,0)$ are the wrenches exerted on the object by the hand, environment, and gravity, respectively. Because the wrench exerted by gravity is constant, it follows that $\vec{w}_{e}= -\vec{w}_{h} + const$. This relation allows us to use the wrenches measured at the hand contact to infer the external contact behavior.

\myparagraph{Measured vs. estimated quantities:} We are able to directly measure the hand pose $(\vec{r}_h, \theta_h)$ and wrench $\vec{w}_e$ via robot proprioception and a force-torque sensor in the wrist of the robot. We have added limited visual feedback, which estimates the world-frame positions of the object vertices $\vec{v}_{i,world}$. Compared to the tactile feedback, vision is relatively noisy and operates at a lower frequency. We primarily use it to seed the estimator with an initial prior, though the estimator is also capable of making use of live visual feedback. Besides the hand pose and wrench, every other quantity, including the object vertices, is estimated.

\subsection{The Constraints of Contact}
\label{subsec:contact_constraints}

\begin{figure}
    \centering
    \includegraphics[width=0.9\columnwidth]{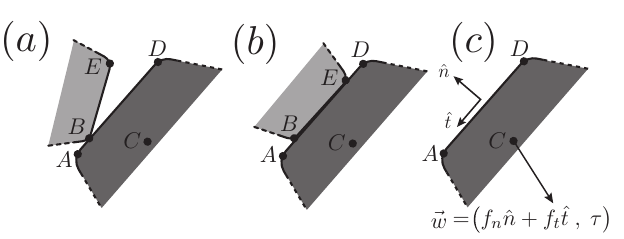}
    \vspace{-0.35cm}
    \caption{a) Point/line contact. b) Flush contact. c) Normal/tangent vectors.}
    \label{fig:contact_mechanics_figure}
    \vspace{-0.5cm}
\end{figure}

Our framework is built on the idea that the constraints of contact can be exploited to regulate the system. Under the quasi-static assumption, these constraints describe the relationship between the contact kinematics and wrenches. 

In our previous work, we reasoned about the friction cone, to detect and regulate sticking/sliding at the hand/environment contact interfaces; and the torque cone, to enforce flush contact between the object and end-effector. These constraints (the generalized friction cone \cite{erdmann1994representation}) are still sufficient to describe the expanded set of interactions, however we must express them in terms of the updated parameterization. We focus on the torque cone, to help reason about more diverse contact geometries.

We consider two contacting bodies, depicted in \figref{fig:contact_mechanics_figure}. Here, $A$ and $D$ are material points fixed to the first body frame, while $B$ and $E$ are material points fixed to the second body frame. The unit vectors $(\hat{n},\hat{t})$ are the normal and tangent of the first body contact face. The wrench $\vec{w} = \left(f_n \hat{n}+f_t\hat{t}, \tau \right)$ is the net contact wrench that the second body exerts on the first body, measured with respect to reference point $C$, which is fixed in the first body frame. We consider point/line contact (\figref{fig:contact_mechanics_figure}a), flush contact (\figref{fig:contact_mechanics_figure}b), and no contact.

\myparagraph{Making and breaking contact:} 
The contact wrench can only be nonzero during contact. During periods of no contact, the contact wrench is zero. For the interaction in \figref{fig:contact_mechanics_figure}:
\begin{align}
\label{eq:contact_condition}
f_n>0 \hspace{1.5 mm} \rightarrow \hspace{1.5 mm} (\vec{r}_B-\vec{r}_A) \cdot \hat{n} = 0\hspace{1.5 mm} 
\text{and/or} \hspace{1 mm}  (\vec{r}_E-\vec{r}_A) \cdot \hat{n} = 0
\end{align}
\myparagraph{Sticking, sliding, and the friction cone:} We use the standard linear complementarity constraints to reason about friction and sticking/slipping. Given friction coefficient $\mu$, and assuming either of the contact geometries shown in \figref{fig:contact_mechanics_figure}: 
\begin{align}
v_{slide} = \frac{d}{dt}((\vec{r}_b&-\vec{r}_a) \cdot \hat{t}) = v_{slide}^+-v_{slide}^-\\
\label{eq:friction_ineq0}
0 \leq -\mu f_n - f_{t} \ \  &\perp \ \  v_{slide}^+ \geq 0\\
\label{eq:friction_ineq1}
0 \leq -\mu f_n + f_{t} \ \  &\perp \ \  v_{slide}^- \geq 0
\end{align}
The $\perp$ symbol is a shorthand, where $0 \leq u \perp w \geq 0$ implies $u \geq 0$, $w \geq 0$, and $uw = 0$. In our implementation, we do not directly estimate the friction coefficient $\mu$, but instead estimate the wrench constraints themselves (see \cite{previous_paper}). 

\myparagraph{Contact geometry and torque constraints:}
We define the resultant contact torque, $\tau_P$, about some reference point $P$:
\begin{align}
\label{eq:resultant_torque}
\tau_P =\tau + (\vec{r}_C-\vec{r}_P)\times \left(f_n \hat{n}+f_t\hat{t}\right)
\end{align}
For the contact geometries shown in \figref{fig:contact_mechanics_figure}, we see that:
\begin{align}
\label{eq:torque_ineq}
\tau_A\le 0,\hspace{1.5 mm} \tau_B = 0,\hspace{1.5 mm} \tau_E\ge 0 \hspace{2 mm} (\text{point/line contact at $B$)}\\
\label{eq:flush_torque_constraints} \tau_A\le 0,\hspace{1.5 mm} \tau_B\le 0,\hspace{1.5 mm} \tau_D\ge 0,\hspace{1.5 mm} \tau_E\ge 0 \hspace{2 mm} \text{(flush contact)} 
\end{align}
These constraints allow us to disambiguate contact geometries, estimate contact locations, and regulate the contact geometry. Here, it is convenient to consider the center of pressure (COP) of a line contact defined by points $G$ and $H$ e.g., the point on the contact patch for which the resultant contact torque is zero:
\begin{align}
\label{eq:COP_def}
\vec{r}_{Q} = COP(\vec{r}_G,\vec{r}_H) \iff 
\tau_Q = 0,\hspace{.5 mm} \vec{r}_{Q} = \gamma \vec{r}_G + (1-\gamma)\vec{r}_H
\end{align}
For the point contact illustrated in \figref{fig:contact_mechanics_figure}a, $B$ is the COP of $AD$. During estimation, we frequently solve \eqref{eq:COP_def} to compute the COP of a potential contact patch, given its endpoints and the measured contact wrench.

\section{Estimation framework}
\label{sec:estimation_framework}

\begin{figure}
    \centering
    \includegraphics[width=0.9\columnwidth]{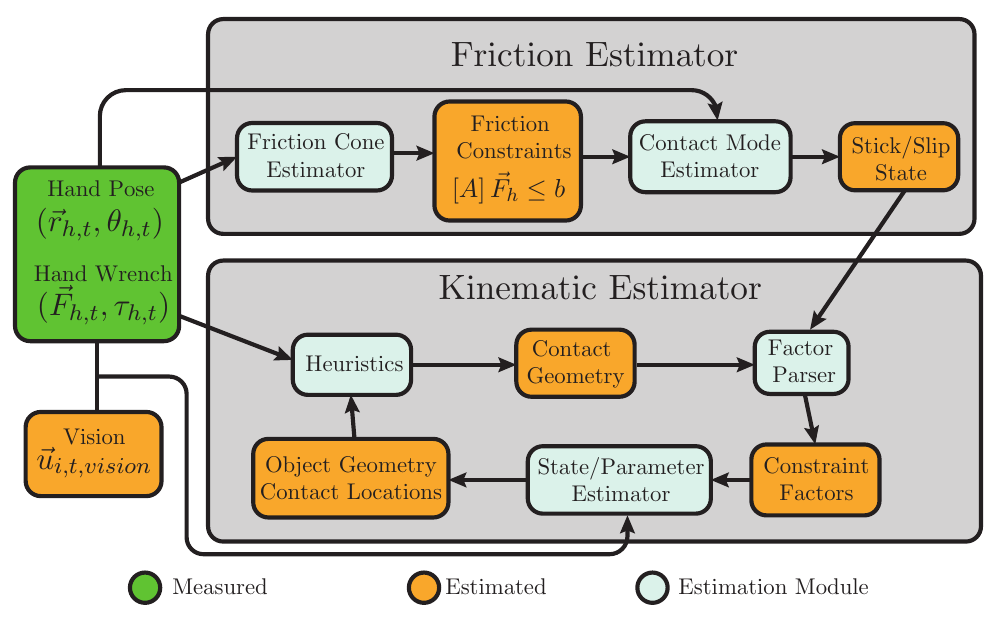}
    \vspace{-0.35cm}
    \caption{The friction and kinematic estimators.}
    \label{fig:estimation_framework_figure}
    \vspace{-0.5cm}
\end{figure}

The estimator consists of two separate modules: the friction estimator and the kinematic estimator (\figref{fig:estimation_framework_figure}). The friction estimator, described in \cite{previous_paper}, builds a model for the friction cones at the hand and ground contact interfaces, which is used to detect and regulate sticking/sliding at both contacts. Our primary focus is the kinematic estimator (\figref{fig:estimation_framework_figure} bottom), which detects the location and geometry of each contact. This process has two stages. First, we use a set of heuristics to guess the contact geometry of each interface. Second, these estimates (and the stick/slip estimates from the friction module) are mapped to a set of kinematic/wrench constraints, which are then added as constraint factors in a factor graph. The graph is then solved to find an estimate of the system state and parameters. We warm start this loop using a prior provided by vision.

\subsection{Contact Geometry Estimation}
\label{subsec:contact_geometry_estimation}
We rely on a sequence of heuristics to identify both the active vertices and contact geometry using tactile feedback and the previous state/parameter estimate. These heuristics are built from the kinematic/wrench constraints laid out in \secref{subsec:contact_constraints} and some geometric reasoning.

\myparagraph{The object/hand interface:} There are four possible contact geometries that we will consider at the object/hand contact interface: no contact, object-line/hand-point, object-point/hand-line and flush contact. We omit object-point/hand-point contact, because it is difficult to maintain and regulate. 

\emph{Contact detection:} The robot and object are assumed to be in contact if the magnitude of the measured force felt at the object/hand contact interface is above a set threshold (around $2-3 \text{N}$). Below this threshold, we assume no contact.

\emph{Object-line/hand-point:} To test for this geometry (\figref{fig:contact_modes_figure}b), we use the measured wrench to compute the COP of the hand contact patch. If this COP is sufficiently close to one of the end-effector boundary points, then object-line/hand-point contact is detected.

\emph{Object-point/hand-line:} For this case (\figref{fig:contact_modes_figure}c), we once again examine the COP of the hand. As per \eqref{eq:torque_ineq}, during object-point/hand-line contact, the hand COP is coincident with the contact vertex. If the estimated positions of the hand COP and one of the object vertices are close, then object-point/hand-line contact is likely. In this case, we then test the kinematic feasibility of this geometry, taking into account the other active contact constraints. A similar feasibility test is performed for flush contact, and the results are compared to determine which geometry is more likely. We supplement this with a heuristic that leverages intuition of how the hand COP changes as a function of the end-effector pose. During flush contact, the COP often fluctuates while the hand remains stationary; whereas during object-point/hand-line contact, the COP motion is constrained by \eqref{eq:torque_ineq}. We use this  to compare the likelihood of these two geometries.

\emph{Flush contact:} If contact has been detected, but the previous two geometries have been ruled out as possibilities, then we assume that the system is in flush contact (\figref{fig:contact_modes_figure}a). We then identify the current contact face by comparing the surface normal of the end-effector to the outward facing surface normals of the object in the most recent state estimate.

\myparagraph{The object/environment contact(s):} Here, we assume that the object is always in contact with the ground. Ground contact is limited to either object-point/ground-line or flush contact. We also check to see if wall contact is occurring. 

\emph{Object-point/ground-line vs. flush contact:} Given the most recent kinematic estimate and tactile feedback, we perform a sequence of tests. First, if one of the object vertices is sufficiently below the rest, then we assume single point contact with the ground at that vertex (\figref{fig:contact_modes_figure}d). Otherwise, if the two lowest object vertices have similar altitude e.g., the bottom edge of the object polygon is near horizontal, we compute the COP of that edge (assuming a weightless object). If the COP is in the interior of this edge by some margin, then we assume that the edge is in flush contact with the ground (\figref{fig:contact_modes_figure}e). Otherwise, the closest vertex to the COP is in single point contact with the ground.

\emph{Wall contact:} To determine whether or not the object is in contact with a wall (\figref{fig:contact_modes_figure}f), we compare the measured force at the hand contact with the friction constraints that we estimate for the ground contact \cite{previous_paper}. These estimated constraints  are computed during periods when there are no external contacts besides the ground; after which they are frozen in place (during periods when wall contact is allowed). Wall contact is detected when the measured contact force significantly violates one of the friction constraints for ground contact, since it implies the existence of a new contact that is generating the wrench causing the constraint violation.

\subsection{Object and Contact Localization}
\label{subsec:object_and_contact_localization}
We now synthesize the contact mode/geometry estimates with tactile feedback to estimate the contact locations. 

\begin{figure}
    \centering
    \includegraphics[width=0.83\columnwidth]{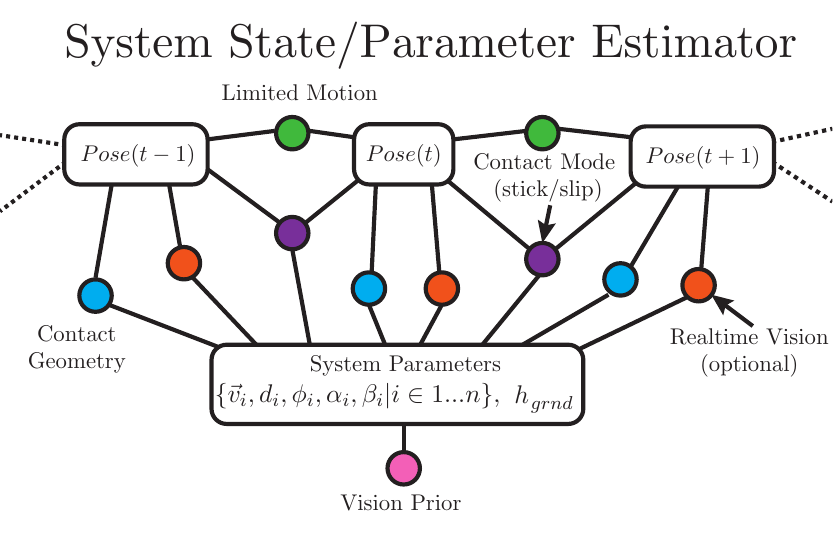}
    \vspace{-0.35cm}
    \caption{Factor graph used for the kinematic estimator. Different sets of contact mode and geometry factors are used at each time-step, with the composition depending upon the stick/slip and contact geometry estimates.}
    \label{fig:factor_graph_figure}
    \vspace{-0.5cm}
\end{figure}

\myparagraph{Factor graphs and GTSAM:} We use factor graphs to formulate the problem of localizing the object and contacts (Tactile SLAM \cite{suddhu2021SLAM}). Factor graphs are bipartite graphs which encode variables and constraints as vertices, and their input-output relationships as edges. For our purposes, factor graphs provide a straightforward way to leverage our knowledge of the contact constraints in order to localize the object and contacts (\figref{fig:factor_graph_figure}). Here, the contact geometry factors (depicted in blue) constrain the instantaneous contact pose and wrench, while the contact mode factors (purple) constrain how the contact pose evolves over time. Each factor is expressed as a function that maps the estimated variables to a constraint violation error. Each constraint has an associated variance that reflects our confidence in its accuracy. There are various algorithms available for computing the optimal values of the estimated variables. We rely on the factor graph software GTSAM \cite{gtsam}, and its implementation of the incremental smoothing and mapping algorithm \cite{ISAM2}.

\myparagraph{Variable factors and measured quantities:} The following are variables in our factor graph:
\begin{itemize}
\item The object-frame vertex positions, $\vec{v}_i = \left(x_i,y_i\right)$.
\item The parameters describing the object-frame surface normals $\phi_i$ and face offsets $d_i$.
\item The object pose in the world-frame, $(\vec{r}_{o,t},\theta_{o,t})$.
\item During flush contact, the tangential displacement between the hand and the object, $s_{t}$.
\item The height of the ground $h_{grnd}$.
\item The gravitational torque parameters $(\alpha_i,\beta_i)$.
\end{itemize}
We use F/T sensing and robot proprioception to measure:
\begin{itemize}
\item The hand contact wrench $\vec{w}_{h,t} = (\vec{F}_{h,t},\tau_{h,t})$.
\item The hand pose $(\vec{r}_{h,t},\theta_{h,t})$.
\end{itemize}
We also assume that the ground is horizontal, meaning that $\hat{n}_g$ and $\hat{t}_g$ are respectively vertical and horizontal.

When describing the constraint factors in this paper, we let $\vec{u}_{i,t}$ denote the current world-frame positions of the object vertices. These are computed as a function of the aforementioned estimation variables, specifically the object pose and the object-frame vertex positions:
\begin{align}
\vec{u}_{i,t} = R(\theta_{o,t})\vec{v}_i+\vec{r}_{o,t} 
\end{align}
where $R(\theta)$ is the rotation matrix for angle $\theta$. When the hand is in flush contact with object face $c$, we use:
\begin{align}
\label{eq:flush_contact_equation}\vec{u}_{i,t} = R(\theta_{h,t}-\phi_c+\pi)\vec{v}_i+d_c\hat{n}_{h,t}+s_t\hat{t}_{h,t} 
\end{align}

The estimator may be provided with noisy live visual feedback of the world-frame vertex positions, $\vec{u}_{i,t,vision}$.

\myparagraph{Constraint factors:} We convert the estimated contact mode and geometry into a set of constraint factors that are added to the factor graph at each time-step.

\emph{Contact:} The end-effector intersects any object point(s) it is in contact with. The same holds true for the ground:
\begin{align}
\left(\vec{u}_{i,t}-\vec{r}_{h,t}\right)\cdot \hat{n}_{h,t} = 0 &| i \in \{\text{hand contact vertices}\} \\
\vec{u}_{i,t}\cdot\hat{n}_g - h_{grnd} = 0 &| i \in \{\text{ground contact vertices}\}
\end{align}
During flush contact with the hand, the hand contact constraint is implicit to \eqref{eq:flush_contact_equation}, so we do not include a factor.

\emph{Sticking contact:} 
If the object is in sticking contact with the hand, then the tangential motion (relative to the hand) of the contact point(s) is zero:
\begin{align}
\left(\vec{u}_{i,t+1}-\vec{r}_{h,t+1}\right)\cdot\hat{t}_{h,t+1}-\left(\vec{u}_{i,t}-\vec{r}_{h,t}\right)\cdot\hat{t}_{h,t} = 0\\
s_{t+1}-s_{t} = 0 \hspace{2 mm} \text{(flush contact)}
\end{align}
Similarly, if the object is in sticking contact with the ground, the horizontal motion of the ground contact(s) is zero:
\begin{align}
\left(\vec{u}_{i,t+1}-\vec{u}_{i,t}\right)\cdot \hat{t}_g = 0
\end{align}

\emph{Torque Balance:} When the object is in single point contact with ground, but not in wall contact, then by the quasi-static assumption, the net torque about the ground contact is zero. 
\begin{align}
\left(\vec{r}_{h,t}-\vec{u}_{i,t}\right)\times \vec{F}_{h,t} + \tau_{h,t} + \alpha_i \cos{\theta_{o,t}}+ \beta_i \sin{\theta_{o,t}} = 0
\end{align}
Here, the quantity $\alpha_i \cos{\theta_{o,t}}+ \beta_i \sin{\theta_{o,t}}$ represents the gravitational torque about the ground contact vertex.

When the object is in single point contact with the hand, then the object contact vertex should be coincident with the COP of the hand contact patch, which we can compute by applying \eqref{eq:COP_def} to the hand pose and measured contact wrench:
\begin{align}
\vec{u}_{i,t}-\vec{r}_{COP,h,t} = 0, \hspace{2 mm} i = \text{hand contact vertex}
\end{align}

\emph{Vision:} Any vision estimates are added as extra constraints:
\begin{align}
\label{eq:vision_constraints}
\vec{u}_{i,t}-\vec{u}_{i,t,vision} = 0
\end{align}

\emph{Geometric consistency:} These constraints ensure that object surface normal $\phi_i$ and face offset $d_i$ parameters are geometrically consistent with the object vertices $\vec{v}_i$:
\begin{align}
\label{eq:geometric_consistency}
d_i = \hat{n}_i\cdot\vec{v}_i &\hspace{1 mm} \rightarrow \hspace{1 mm} x_i\cos\phi_i+y_i\sin\phi_i-d_i = 0\\
d_i = \hat{n}_i\cdot\vec{v}_{i+1} &\hspace{1 mm} \rightarrow \hspace{1 mm} x_{i+1}\cos\phi_i+y_{i+1}\sin\phi_i-d_i = 0
\end{align}

\emph{Regularization constraints:} We include low-weight constraint factors to limit the object motion between time-steps. 
\begin{align}
\vec{r}_{o,t+1}-\vec{r}_{o} = 0, \hspace{2 mm} \theta_{o,t+1}-\theta_{o,t} = 0\\
s_{t+1}-s_{t} = 0 \hspace{2 mm} \text{(flush contact)}
\end{align}
\myparagraph{Optimization weights:} Each constraint factor has an associated variance that reflects our confidence in its accuracy. This is essentially an optimization weight, with lower variances increasing the significance of their corresponding factor in the optimization. We tuned these variances manually.

\subsection{Estimator Initialization and the Role of Vision}
\label{subsec:estimator_initialization}
The contact geometry heuristics (\secref{subsec:contact_geometry_estimation}) and the factor graph estimator (\secref{subsec:object_and_contact_localization}) are mutually dependent: the contact geometry estimator needs a somewhat accurate state/parameter estimate to disambiguate contact geometries through tactile feedback, and the factor graph based estimator relies on the contact geometry estimator to determine which constraints are imposed at each time-step. This presents a chicken and egg problem: if the initial estimation error is too large, it will never correctly converge.

We resolve this by using vision to generate a prior of the object's geometry and starting pose to warm start the kinematic estimator. Once it has been initialized, the kinematic estimator is capable of operating on tactile feedback alone. Live visual feedback is still useful when available, and acts to regularize the object pose. This safeguards against errors from accumulating when the heuristics incorrectly guess the contact geometries.

The vision heuristic we use directly estimates the positions of the object vertices in the world-frame, under the assumption that the object is a convex polygon. We start with a few seed pixels that belong to the object's interior, and assume that the object is a somewhat uniform color that is distinct from its surroundings. With these seed pixels, we perform a breadth-first-search on a down-sampled version of the image to sparsely flood fill the portion of the object in the image, giving us a blob that approximately covers the object. Going back to the original high-resolution image, we then randomly sample pixels near the boundary of this blob to check if they should belong to the object's interior. This gives us a more accurate picture of the object's boundaries. Finally, we extract the object vertices using a heuristic to approximate the 2D convex hull for a noisy data-set (the same method we use to compute the friction constraints from the measured contact wrenches in \cite{previous_paper}). This heuristic was sufficiently reliable for developing the kinematic estimator, but should be replaced with a standard perception method during real-world implementation.

 \begin{figure}
    \centering
    \includegraphics[width=0.9\columnwidth]{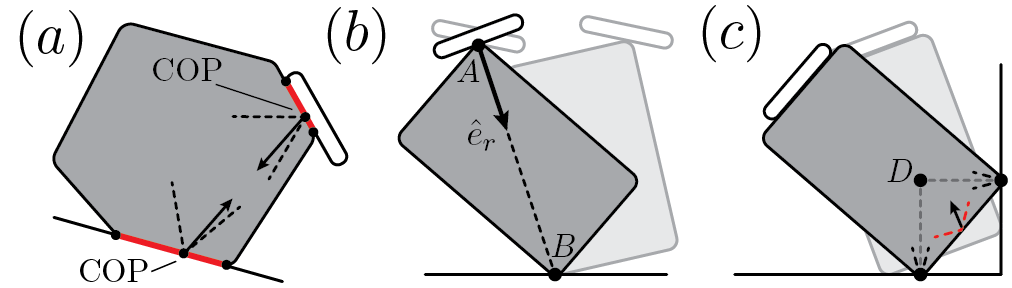}
    \vspace{-0.35cm}
    \caption{The controller relies on the kinematic estimates to regulate the system: a) The controller enforces flush contact at an interface by keeping the COP on the interior of the \textbf{estimated} contact patch. b) During corner/corner pivoting, the admissible motions are pure rotations about the respective contact points ($A,B$), and the constraint direction corresponds to pure translation in the radial ($\hat{e}_r$) direction. c) During wall pivoting, the admissible motion direction corresponds to pure rotation about the ICR, $D$, which is a function of the external contact locations. In addition, the external friction cone (red) is expanded due to the wall contact}
    \label{fig:torque_constraint_figure}
    \vspace{-0.5cm}
\end{figure}

\section{Control framework}
\label{sec:control_framework}
We further develop the control framework that we presented in \cite{previous_paper}. In this framework, the contact configuration controller is built on top of a lower level impedance controller that regulates the end-effector pose. At each time-step, the controller solves a quadratic program (QP) to determine the incremental change of the impedance target, $\Delta \vec{x}_{tar}$:
\begin{align}
\label{eq:example_qp_cost}
& \min_{\Delta \vec{w}_h, \Delta \vec{x}_{tar}} \alpha_0  || \Delta \vec{x}_{tar} ||^2 + \sum \alpha_i \big(\Delta \vec{x}_{tar} \cdot \Delta\vec{q}_i - \beta_i \Delta \epsilon_i \big)^2\\
\label{eq:example_qp_iq_const}
 & \ \ \ s.t. \quad \ \  \hat{n}_j\cdot (\vec{w}_{h,meas} +  \gamma_j \Delta  \vec{w}_h) \leq b_j \quad \forall j \\
 \label{eq:example_qp_eq_const}
& \qquad \qquad \Delta \vec{w}_h =  K \Delta\vec{x}_{tar}
 \end{align}
 Here, the sum $\vec{w}_{meas} + \Delta  \vec{w}_h$ is the predicted wrench that the hand will exert after the impedance target has been incremented by $\Delta\vec{x}_{tar}$. The constraints \eqref{eq:example_qp_iq_const} correspond to the estimated value of the wrench space constraints, which are weighted by $\gamma_j$. The equality constraint \eqref{eq:example_qp_eq_const} corresponds to the stiffness law of the low-level impedance controller.  Finally, the cost \eqref{eq:example_qp_cost} is used to regulate the pose along the admissible motion directions of the system, $\Delta\vec{q}_i$. The composition of cost terms and wrench constraints is specified by the motion primitive that the controller has been commanded to execute. The main change to this control framework is the adjustment/addition of individual constraints and cost terms, as well as the addition of a few new primitives; however, the underlying QP remains the unchanged.

\myparagraph{Contact Geometry Regulation:} In our previous work, the controller enforced flush contact with the hand via. the inclusion of two torque constraints in the QP:
\begin{align}
\label{eq:torque_regulation0}
 -l_h (\vec{F}_{h,meas}+ \Delta  \vec{F}_h ) \cdot \hat{n}_h - (\tau_{h,meas} + \Delta  \tau_h) \le -\epsilon \\ 
 \label{eq:torque_regulation1}
 l_h (\vec{F}_{h,meas}+ \Delta  \vec{F}_h ) \cdot \hat{n}_h + (\tau_{h,meas} + \Delta  \tau_h) \le -\epsilon
 \end{align}
This ensured that the COP of the object/hand contact remained in the interior of the hand by some safety margin $\epsilon$. These constraints rely on the assumption that the boundaries of the object contact face extends past the hand, and will fail in the cases of overhang (\figref{fig:torque_constraint_figure}a). Since the improved estimator keeps track of the object vertices at the hand contact, we can adjust the torque constraints as follows:
\begin{align}
 -l_a (\vec{F}_{h,meas}+ \Delta  \vec{F}_h ) \cdot \hat{n}_h - (\tau_{h,meas} + \Delta  \tau_h) \le -\epsilon \\ 
 l_b (\vec{F}_{h,meas}+ \Delta  \vec{F}_h ) \cdot \hat{n}_h + (\tau_{h,meas} + \Delta  \tau_h) \le -\epsilon
 \end{align}
The constants $l_a$ and $l_b$ take into account the possibility of overhang; and are a function of the estimated object vertex positions, and the hand's pose and length. 

We have also constructed an analogous set of torque constraints that enforce flush contact with the environment. With these constraints, we have created variants of the three original motion primitives (\figref{fig:contact_transitions_figure} a-c). These new primitives can slide the object or hand while maintaining flush external contact, or rotate an object to force a transition from single point external contact to flush contact  (\figref{fig:contact_transitions_figure}d).

 \begin{figure}
    \centering
    \includegraphics[width=0.9\columnwidth]{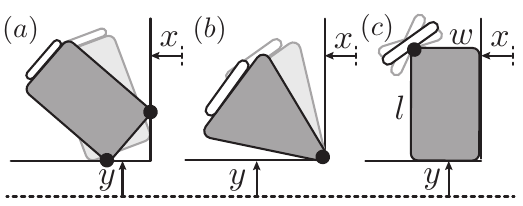}
    \vspace{-0.35cm}
    \caption{Experimental design: we manipulate an object while enforcing wall contact, which provides ground truth for measuring the estimation error. a) Pivoting a rectangle against a corner, and computing the estimation error of the y-coordinate of the bottom vertex and the x-coordinate of the wall contact vertex. b) Pivoting a triangle against a corner, and computing the estimation error of the pivot vertex. c) Pivoting about the vertex of a rectangle that is in flush contact with a wall, and computing the estimation error of the pivot vertex.}
    \label{fig:experiments_figure}
\end{figure}

\myparagraph{Corner/corner pivoting:} Corner/corner pivoting is a new primitive during which the object is in sticking single point contact with both the ground and the hand (\figref{fig:torque_constraint_figure}b). Here, the admissible motion directions $\Delta\vec{q}_\phi = \left(R(90^\circ)(\vec{r}_h-\vec{r}_A) ,1\right)$ and $\Delta\vec{q}_\theta=\left(R(90^\circ)(\vec{r}_h-\vec{r}_B) ,1\right)$ are rotations about the hand contact and external contact respectively; and the constraint direction $\hat{n}_r = (\hat{e}_r,0)$ corresponds to pure translation along the line segment connecting the two contacts. We build a new QP using the cost terms and constraints corresponding to $(\Delta\vec{q}_\phi,\Delta\vec{q}_\theta,\hat{n}_r)$, which gives us a controller that regulates the system during this contact configuration.

\myparagraph{Wall pivoting:} Finally, we have constructed a primitive that regulates the orientation of an object, while maintaining sticking flush contact with the hand and sliding contact at both external contacts (\figref{fig:torque_constraint_figure}c). This is a variant of our original rotation primitive (\figref{fig:contact_transitions_figure}a), with two key changes. First, the admissible motion direction for rotation, $\Delta\vec{q}_\theta = \left(R(90^\circ)(\vec{r}_h-\vec{r}_D) ,1\right)$, takes into account the adjusted instantaneous center of rotation (ICR) location due to wall contact. Second, the external friction constraints \eqref{eq:example_qp_iq_const} take into account the fact that the wall contact expands the environment friction cone, since it can generate an arbitrary amount of force in the direction normal to the wall.

\section{Experiments}
\label{sec:experiments}
We perform a set of experiments to validate the estimator. In our setup, we manipulate an object while enforcing contact with a wall (\figref{fig:experiments_figure}). Under this contact constraint, the known locations of the ground ($y$), wall ($x$), and the object dimensions ($w,l$) act as ground truth for specific coordinates subsets of the contact locations. The composition of those subsets is determined the object and the type of wall contact. We compare the root mean square error (RMSE) of these coordinate subsets when computed by our vision heuristic (V only), the kinematic estimator making use of live visual feedback (K+V), and the kinematic estimator using a vision prior, but without live visual feedback (K only). The results (Table \ref{table:experiment_results}) indicate that F/T sensing and robot proprioception are often sufficient for estimating the geometry and locations of the external contacts, which is especially important in situations where the system does not have constant access to visual feedback (object or wall occlusions).

 \begin{figure}
    \centering
    \includegraphics[width=0.9\columnwidth]{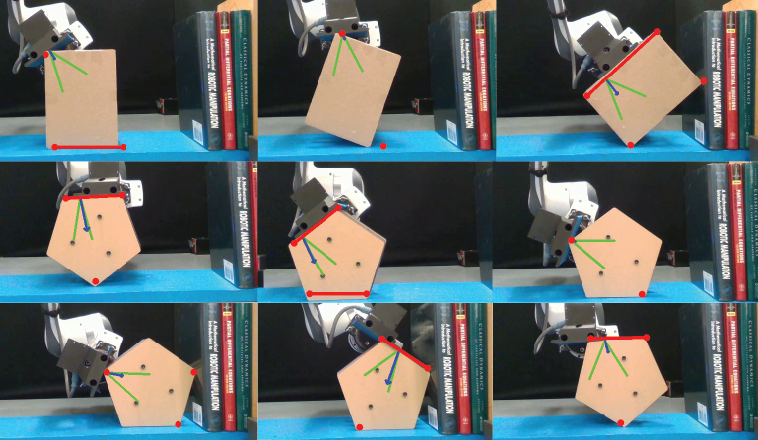}
    \vspace{-0.35cm}
    \caption{Demonstrations: To test our framework, we regulate the system through a sequence of contact configurations. The measured robot wrench (blue) and the robot pose are used to estimate the friction constraints (green) and the contact locations and geometries (red). \textbf{Top row:} We execute the corner/corner pivoting primitive to move a rectangle from line contact with the ground to wall contact. \textbf{Bottom rows:} We regulate a pentagon through sequence of contact geometry combinations.}
    \label{fig:demonstration_figure}
\end{figure}

\begin{table}
\caption{\label{table:experiment_results}}
\vspace{-1 mm}
\begin{tabular}{ |c|c|c|c| } 
 \hline
 Trial & V only & K + V & K only \\
 \hline
 (a) Wall Pivot &  3.28 (cm)	 & 1.68 (cm)	  & 1.66 (cm) \\
 \hline
 (b) Triangle Pivot & .74 (cm)	 & .62 (cm)	 & 1.00 (cm) \\
 \hline
 (c) Object Corner & .38 (cm) & .53 (cm)	 & .34 (cm) \\
 \hline
\end{tabular}
\end{table}

\section{Demonstrations}
\label{sec:demonstrations}
To demonstrate our framework, we use it to regulate various shapes through a predetermined sequence of contact geometries (\figref{fig:demonstration_figure}). For these tests, the kinematic estimator only relies on tactile feedback (after the initial vision prior). We manually jog the hand to disengage from a face/vertex of the object and then reengage at a different face/vertex.

\section{Conclusion}
\label{sec:conclusion}
In conclusion, we show that, with a reasonably accurate prior, it is possible to use tactile feedback to infer complex contact configurations between the object and the hand/environment. This can then be used to regulate the system as it transitions through these configurations. 

One of the biggest challenges of this work was developing the heuristic that differentiates object-point/hand-line contact from flush hand contact, using tactile feedback alone. Though it often works, a momentary failure can be disastrous. Since the kinematically feasible object poses for the two geometries can differ significantly, a bad guess can cause the kinematic estimate to permanently diverge from the correct solution. Though we plan to improve this heuristic, this issue demonstrates how live visual feedback synergizes with tactile estimation: even noisy low-frequency visual estimates can be enough to regularize the object pose, allowing the estimator to recover when a faulty contact geometry guess was made.

In the future, we plan on designing an additional estimator that predicts the feasibility of various motions and discrete transitions of the system using the kinematic and friction estimates. This would then be combined with a high-level planner that would sequence primitives, both to excite the system for the purpose of estimation, and to drive an object to a target state.

\bibliographystyle{IEEEtran}
\bibliography{references}

\end{document}